\documentclass[10pt,twocolumn,letterpaper]{article}

\usepackage{cvpr}
\usepackage{times}
\usepackage{epsfig}
\usepackage{graphicx}
\usepackage{amsmath}
\usepackage{amssymb}
\usepackage{pdfpages}
\usepackage{multirow}
\usepackage{booktabs}
\usepackage{subcaption}
\usepackage{bm}
\usepackage{soul}

\newcommand*{\affmark}[1][*]{\textsuperscript{#1}}

\usepackage[breaklinks=true,bookmarks=false]{hyperref}

\cvprfinalcopy 

\def\cvprPaperID{****} 

\begin{document}

\title{Locality Aware Appearance Metric for Multi-Target Multi-Camera Tracking}

\author{Yunzhong Hou \affmark[1] ~~~ Liang Zheng \affmark[1] ~~~ Zhongdao Wang \affmark[2] ~~~ Shengjin Wang \affmark[2]\\
\affmark[1] Australian National University ~~~ \affmark[2] Tsinghua University\\
{\tt\small \affmark[1]\{firstname.lastname\}@anu.edu.au ~~~ \affmark[2]\{wcd17@mails.tsinghua.edu.cn, wgsgj@tsinghua.edu.cn\}}
}


\maketitle


\begin{abstract}

Multi-target multi-camera tracking (MTMCT) systems track targets across cameras. Due to the continuity of target trajectories, tracking systems usually restrict their data association within a local neighborhood. In single camera tracking, local neighborhood refers to consecutive frames; in multi-camera tracking, it refers to neighboring cameras that the target may appear successively. 
For similarity estimation, tracking systems often adopt appearance features learned from the re-identification (re-ID) perspective. Different from tracking, re-ID usually does not have access to the trajectory cues that can limit the search space to a local neighborhood. Due to its global matching property, the re-ID perspective requires to learn global appearance features. We argue that the mismatch between the local matching procedure in tracking and the global nature of re-ID appearance features may compromise MTMCT performance.


To fit the local matching procedure in MTMCT, 
in this work, we introduce locality aware appearance metric (LAAM). Specifically, we design an intra-camera metric for single camera tracking, and an inter-camera metric for multi-camera tracking. Both metrics are trained with data pairs sampled from their corresponding local neighborhoods, as opposed to global sampling in the re-ID perspective.
We show that the locally learned metrics can be successfully applied on top of several globally learned re-ID features. 
With the proposed method, we report new state-of-the-art performance on the DukeMTMC dataset, and a substantial improvement on the CityFlow dataset.

\end{abstract}

\section{Introduction}
\label{intro}

\begin{figure}
    \centering
    \includegraphics[width=\linewidth]{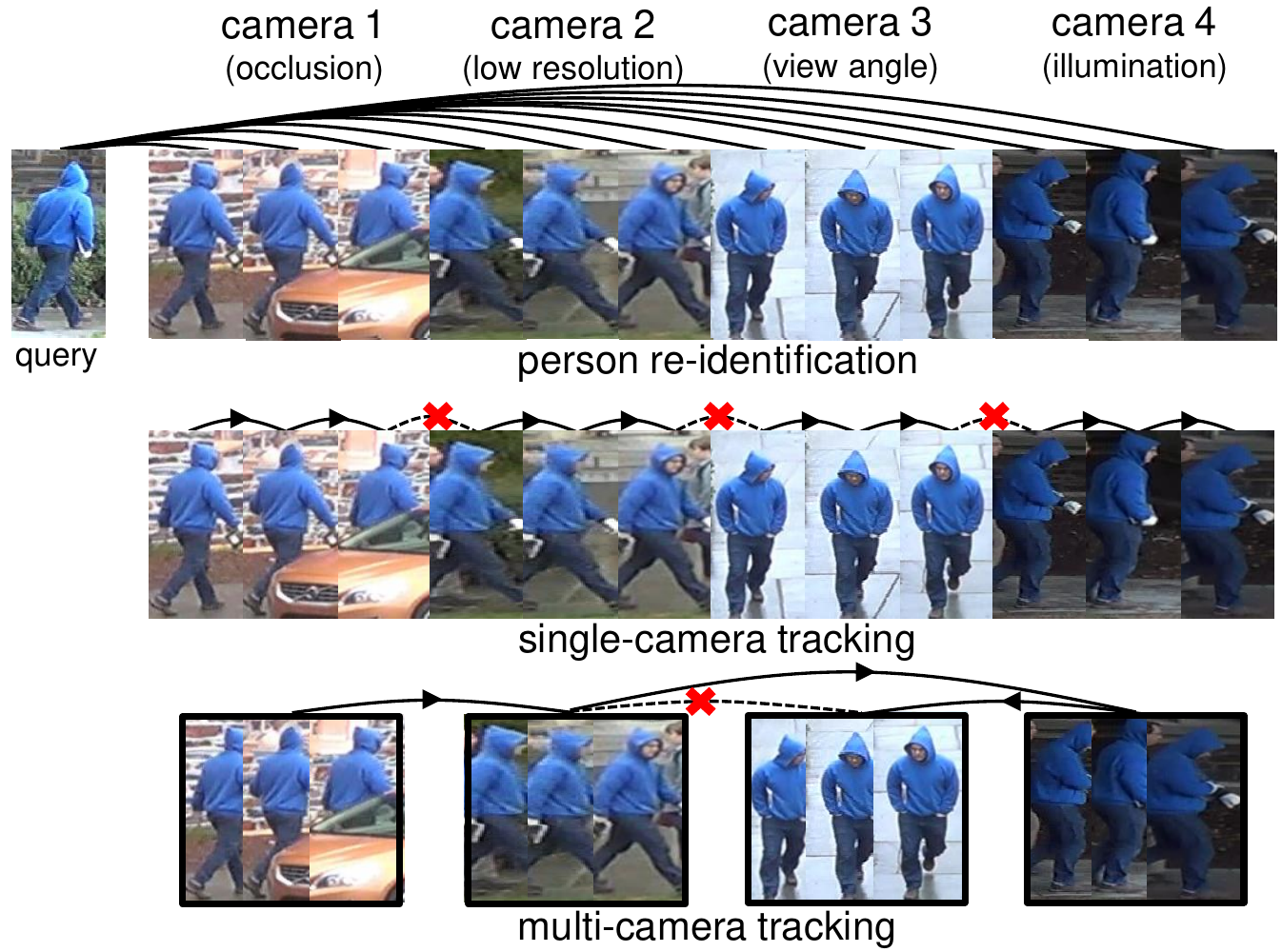}
    \vspace{-6mm}
    \caption{Difference between re-ID and MTMCT. Given a query, re-ID searches the \emph{global gallery} for true matches from all cameras. 
    In comparison, MTMCT matches within a local neighborhood, in single camera tracking (SCT) and multi-camera tracking (MCT). 
    Specifically, in MCT, when the target is in camera 2, we do not consider camera 3, since targets never appear in these two cameras successively (cameras may be too far away).  
    }
    \vspace{-3mm}
    \label{fig:reid_vs_tracking}
\end{figure}

\begin{figure*}
    \centering
    \includegraphics[width=0.95\linewidth]{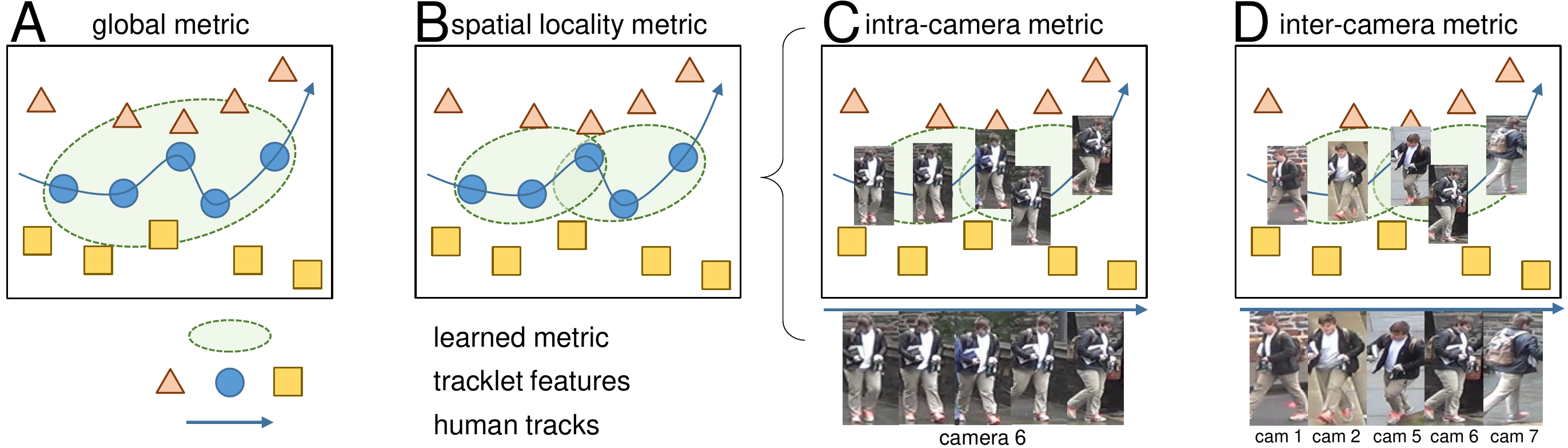}
    \caption{\textbf{(A)} A global metric learned from the entire train set considers all data. This metric is relatively robust but has a slack decision boundary (errors often exist). 
    \textbf{(B)} This paper proposes a locally learned metric. It has a tight decision boundary and is more sensitive. In MTMCT, data association is usually within a local neighborhood, as opposed to global matching in re-ID. So local metric learning suits better. 
    The proposed locality aware appearance metric (LAAM) has \textbf{(C)} an intra-camera metric for SCT and \textbf{(D)} an inter-camera metric for MCT.
    The former is learned on tracklet pairs within a time period in the same camera. The latter is learned on tracklet pairs across neighboring cameras (that the target may appear successively). 
    }
    \vspace{-2mm}
    \label{fig:TLML_intro}
\end{figure*}

Multi-target multi-camera tracking (MTMCT) aims to identify and locate all targets in a multi-camera system at all times. 
In a closely related task, given a probe image, re-identification (re-ID) systems search the gallery to retrieve images of the same identity. 


An MTMCT system is composed of several components, including detection, similarity estimation, and data association. 
Based on similarity estimations, detected object bounding boxes are associated in first single camera tracking (SCT), and then multi-camera tracking (MCT).
Since the targets, \eg, pedestrians, vehicles, have continuous trajectories, most tracking systems only search the \textit{local neighborhood} for data association. For example, temporal sliding window techniques are employed in many MTMCT systems~\cite{ristani2018features,ristani2016performance,tesfaye2017multi}. In SCT, temporal sliding windows restrict the local matching neighborhood to the consecutive frames within a camera. In MCT, these windows restrict the local matching neighborhood to the neighboring cameras that the target may appear successively. 

The appearance feature is a driving force in MTMCT. Currently, the tracking community shares very similar appearance representations and deep learning architectures with the re-ID community. That is, the feature is learned \textit{globally} from the entire train set, and then applied to both SCT and MCT \cite{ristani2018features,zhang2017multi}. 

However, MTMCT and re-ID have their differences. 
First, Re-ID systems usually do not have access to trajectories, camera topology, and other spatial-temporal cues. 
Second, SCT is a very important part of MTMCT. In contrast, re-ID ignores candidates of the same camera as the probe in its evaluation~\cite{zheng2015scalable}. 
%

In this study, we further explore the third and most important difference: local matching in tracking versus global matching in re-ID. 
As shown in Fig.~\ref{fig:reid_vs_tracking}, 
MTMCT only associates data within within a \textit{local neighborhood} (smaller appearance variances). 
Specifically, in SCT, only consecutive frames within single camera video are searched. In MCT, we match in a limited camera pool, as the search scope is narrowed by temporal sliding windows. 
On the other hand, 
given a query image, re-ID searches a \textit{global gallery} covering all cameras (large appearance variances). 
This \textit{local versus global} difference is non-trivial. When applying the re-ID features directly, the \textit{mismatch} between local matching in tracking and global re-ID appearance feature may compromise the MTMCT performance. 

In fact, we believe this local versus global mismatch is the reason for the phenomenon Ristani \etal noticed. In their work~\cite{ristani2018features}, 
it is found that high-performing re-ID features do not necessarily lead to good MTMCT performance. 
In fact, re-ID models learn to deal with \textit{all} kinds of environmental variances. However, in SCT, we only need to match consecutive frames that have relatively small (compared to cross-camera) appearance changes. In MCT, we still do not need to consider all environmental variance. For example, features for MCT do not need to be robust against viewpoint variance and low-resolution simultaneously (Fig. \ref{fig:reid_vs_tracking}), since targets never appear in these cameras successively. 
In such cases, a stronger re-ID appearance feature does not necessarily lead to a higher MTMCT result. 

To fit the local matching procedure in tracking, this paper proposes a locality aware appearance metric (LAAM). 
Specifically, for SCT, we sample training data pairs from consecutive frames within a single camera. For MCT, the training data pairs are selected from neighboring cameras (that the target may appear successively). 
Using two sampling strategies, we have an intra-camera metric for SCT and an inter-camera metric for MCT (see Fig.~\ref{fig:TLML_intro}). 

We show that LAAM can effectively improve tracking accuracy on multiple datasets, including a pedestrian dataset, DukeMTMC~\cite{ristani2016performance}, and a vehicle dataset, CityFlow~\cite{tang2019cityflow}. It can also be applied and on top of multiple re-ID features, such as IDE \cite{zheng2016person}, PCB \cite{sun2018PCB} and the triplet feature \cite{hermans2017defense}. With a competitive tracker \cite{ristani2018features}, we report the state-of-the-art accuracy on the DukeMTMC dataset.

\begin{figure*}[ht]
    \centering
    \includegraphics[width=0.95\linewidth]{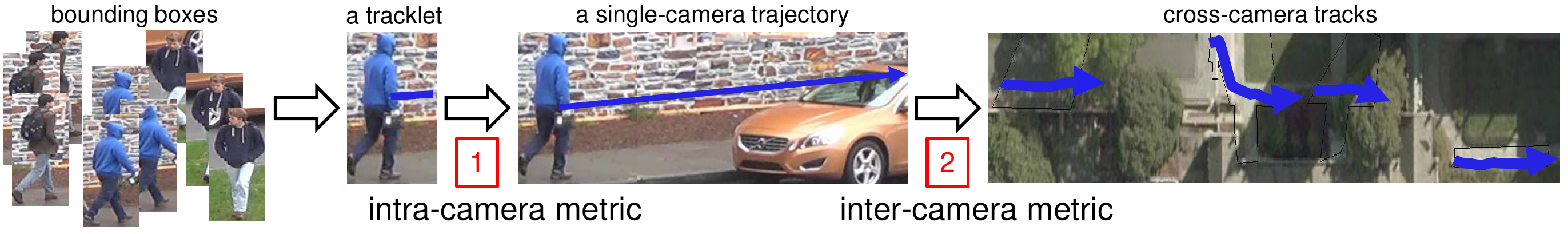}
    \vspace{-1mm}
    \caption{MTMCT system overview. Given object bounding boxes, we first connect the bounding boxes into short but reliable \emph{tracklets}. Then, the tracklets are merged into \emph{single camera trajectories}. Finally, single camera trajectories are associated to form \emph{cross-camera tracks}. 
    The proposed LAAM includes an intra-camera metric and a inter-camera metric. Intra/inter camera metrics are applied to generate \emph{single camera trajectories} and \emph{cross-camera tracks}, respectively.} 
    \label{fig:system}
    \vspace{-4mm}
\end{figure*}


\section{Related Work}\label{related work}

\textbf{Multi-object tracking in a single camera.}
In MTMCT, the SCT step is inspired by multi-object tracking (MOT) ~\cite{leal2015motchallenge,milan2016mot16,leal2017tracking}. There are both online and offline methods. Online tracking methods should not use data from the future time slots. They usually associate detections to tracklets in a greedy manner~\cite{fagot2016improving,choi2015near}.
Offline methods can benefit from future information. They usually formulate the problem as batch optimization, such as 
shortest path \cite{berclaz2011multiple,dehghan2015target,wang2016tracking,choi2015near}, bipartite graph~\cite{brendel2011multiobject,cai2014exploring}, and pairwise terms~\cite{leibe2007coupled,wang2014tracklet,hamid2015joint,yu2016solution,joo2007multiple, kumar2014multiple,chari2015pairwise,collins2012multitarget,das2014consistent,dehghan2015gmmcp}. To reduce computation complexity, some employ a hierarchical approach~\cite{joo2007multiple,singh2008pedestrian,shitrit2014multi}, or temporal sliding windows~\cite{sadeghian2017tracking,milan2014continuous,choi2015near}. 

\textbf{Cross-camera tracking in MTMCT.}
Cross-camera tracking is a unique feature of MTMCT. \cite{tesfaye2017multi,maksai2017non,ristani2018features,yoon2018multiple,zhang2017multi,jiang2018online}. 
On the other hand, offline methods \cite{ristani2018features,ristani2016performance,tesfaye2017multi,zhang2017multi} usually employ batch optimization techniques for higher accuracy, which is similar to MOT trackers. 
Vehicle MTMCT is also studied. In~\cite{tang2018single}, Tang \etal use multiple cues to accommodate the similar appearance, heavy occlusion, and large viewing angle variation in vehicle tracking.

\textbf{Re-ID features and its application in MTMCT.}
Re-ID originated from cross-camera tracking \cite{zheng2016person}.
Recently, this area witnessed many competitive CNN structures being proposed \cite{zheng2015scalable,sun2018PCB}. Loss functions and training techniques are studied, such as the contrastive loss~\cite{varior2016gated}, triplet loss~\cite{schroff2015facenet,cheng2016person,liu2017end} and hard negative mining~\cite{hermans2017defense}. Data augmentation methods are explored to enrich the database \cite{barbosa2018looking,zhong2017random}. The advancement in re-ID has been pushing forward the state-of-the-art in MTMCT \cite{ristani2018features,zhang2017multi,jiang2018online}. 
In~\cite{ristani2018features}, Ristani \etal propose a global feature learning method to improve the performance on both re-ID and MTMCT.

\textbf{Metric learning for multi-camera tasks.}  Metric learning algorithm has been studied in re-ID ~\cite{zheng2016person}. 
Besides, metric learning is also investigated in tracking to compute the similarity between observations. Unlike predefined distance metrics, these learned metrics can automatically adapt to a specific scenario and yield higher accuracy~\cite{bellet2013survey}.
For example, Leal-Taix\'e \etal \cite{leal2016learning} train a Siamese network to aggregate pixel values and optic flow. 
In \cite{xiang2018multiple}, Xiang \etal jointly learn a global feature representation and a distance metric for multi-object tracking. In~\cite{thoreau2018improving}, Thoreau \etal learn a Siamese network from re-ID datasets for similarity estimation in online tracking.

Departing from existing works, this paper studies the intrinsic dissimilarities between MTMCT and re-ID. Instead of directly learning a global feature representation/metric, we investigate locality aware appearance metric (LAAM) to meet the local matching in MTMCT data association. 



\section{MTMCT System Overview} \label{overview}

\textbf{Problem formulation.} 
We follow the graph-based problem formulation introduced in \cite{ristani2018features}. 
We represent observations (bounding boxes, tracklets, trajectories)
as nodes, and the similarities between them as weighted edges in a graph $G = \left( V,E\right)$. 
For a pair of nodes $i,j \in V$, $w_{i,j} \in E$ refers to the estimated similarity between them, and $x_{i,j} \in \left\{-1,1\right\}$ indicates whether they are of the same identity. 
The optimization problem is formulated as follows,
\begin{align}
\begin{aligned}
    \max_{x_{i,j}}  & \sum_{\forall i,j \in V}  x_{i,j}w_{i,j}, \\
    \textit{s.t.} \; \;  & x_{i,j} + x_{j,k} \leq 1+x_{i,k}, \forall i,j,k \in V. 
\end{aligned}
\label{eq:optimize}
\end{align}
Eq.~\ref{eq:optimize} maximizes intra-group similarity, minimizes inter-group similarity, and enforces transitivity (two data should be of same identity if both of them share identity with a third data point). In fact, a better performing similarity estimation $w_{i,j}$ will make this optimization problem easier, and improve association accuracy. 

\textbf{Detection.} For DukeMTMC dataset, we adopt the OpenPose~\cite{cao2018openpose} detector following \cite{ristani2018features}. For CityFlow dataset, we use the SSD~\cite{liu2016ssd} detector provided by AI-City 2019 challenge~\cite{shine2019comparative}.

\textbf{Similarity estimation.} 
In the baseline, 
given a pair of CNN features ${\bm{f}_i}$ and $\bm{f}_j$, their appearance similarity score $w_{i,j}$ is computed as,
\begin{equation} \label{eq:w_a}
    w_{i,j} = \frac{\mathit{thres}-d\left(\bm{f}_i, \bm{f}_j \right)}{\mathit{norm}},
\end{equation}
where $d(\cdot,\cdot)$ is a distance metric, and we simply employ Euclidean distance here. $\mathit{thres} = \frac{\mu_n+\mu_p}{2}$, and $\mathit{norm} = \frac{\mu_n-\mu_p}{2}$. $\mu_p$ and $\mu_n$ denote the average feature distance of the same and different identities, respectively.


\textbf{Data association.} Figure \ref{fig:system} depicts the overall data association procedure. 
First, object bounding boxes are connected into tracklets.
Then, the tracklets are matched into single camera trajectories. 
At last, the single camera trajectories are associated to form cross-camera tracks. 

For SCT, we use short temporal sliding window to associate tracklets. For MCT, much longer temporal sliding window is used in data association, due to the long walking time of targets across cameras. 


\begin{figure*}[ht]
    \centering
    \includegraphics[width=0.9\linewidth]{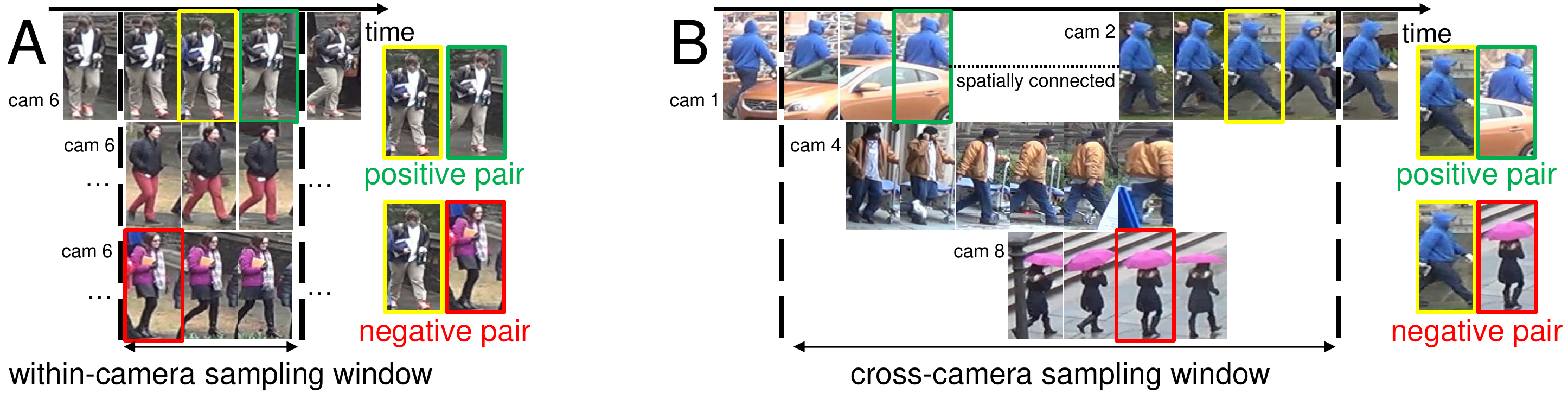}
    \caption{The proposed data sampling strategy for training the intra- and inter-camera metrics. 
    Camera labels of the same person are colored the same. 
    A shorter data sampling window is used to sampled within one camera for the intra-camera metric (\textbf{A}). On the other hand, inter-camera metric uses a longer data sampling window to sample positive data pairs from a different camera, and negative pairs from random cameras (\textbf{B}).}
    \vspace{-2mm}
    \label{fig:TLML_sampling}
\end{figure*}

\begin{figure}[t]
    \centering
    \includegraphics[width=\linewidth]{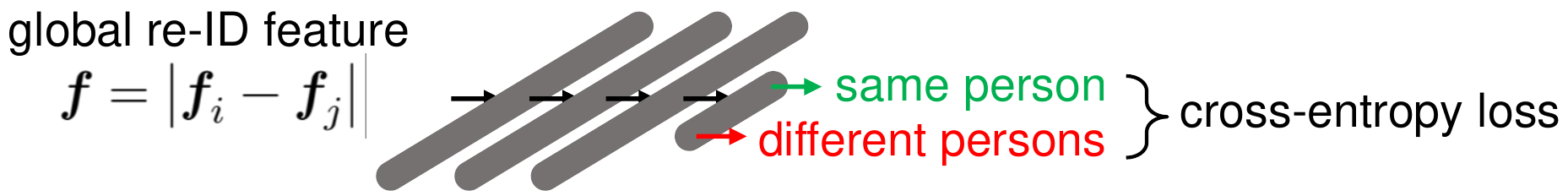}
    \caption{Structure of the metric network for LAAM. It has three fully connected layers and a 2-dim softmax output layer. The network takes the absolute difference vector between a pair of features as input and outputs the confidence score of the input pair belonging to the same person. 
    }
    \vspace{-2mm}
    \label{fig:TLML_sturcture}
\end{figure}

\section{Locality Aware Appearance Metric} \label{TLML}
As mentioned in Section~\ref{overview}, a good  similarity estimation substantially improves association accuracy. 
In this section, we present a novel locality aware appearance metric (LAAM) by focusing on the local neighboring samples. %
Different form re-ID metric that aims at retrieving images from a global gallery, the learned locality aware appearance metric focuses on matching local neighboring candidates, which better fits the local matching tasks in MTMCT.
LAAM is composed of a metric network (not our contribution) and a novel data sampling strategy (main contribution). Descriptions are provided below.

\subsection{Metric Network Structure}\label{structure}
The metric network is used to compute similarity scores between a pair of tracklets or trajectories, both of which are generated by global average pooling bounding box features. In our method, it replaces the Euclidean distance based similarity estimation (Eq.~\ref{eq:w_a}). 

As shown in Fig.~\ref{fig:TLML_sturcture}, the network is a 3-layer perceptron \cite{gardner1998artificial}. The hidden fully connected layers output a $128$-dim vector followed by ReLU activation. 
Given a pair of features $\bm{f}_i$ and $\bm{f}_j$, their absolute difference $\bm{f}=\left|\bm{f}_i - \bm{f}_j\right|$ is used as the network \textbf{input}. 
The \textbf{output} of the metric network is a $2$-dim softmaxed vector, denoted as $\bm{x}=\left(x_0,x_1\right)$. $x_0$ and $x_1$ encode the possibility of the input pair being of different identities or the same identity, respectively. 

During \textbf{training}, the re-ID feature extractor is fixed, and only the metric network is updated. The metric network is trained as a classification problem using a cross-entropy loss function. 
During \textbf{testing}, we exert a scaling factor of $0.1$ onto the softmax layer, to prevent the appearance similarity score from overshadowing other cues, \eg, spatial-temporal cues. The similarity score for the proposed metric is computed by, 
\begin{align}
    w = x_1-x_0. \label{eq:w_metric}
\end{align}
This similarity value should be positive if the data pair belongs to the same identity, and negative if otherwise. 



\subsection{Intra-Camera Metric and Inter-Camera Metric}\label{sec:two_metrics}
LAAM has an intra-camera metric and an inter-camera metric, for SCT and MCT, respectively (Fig. \ref{fig:TLML_intro}). Both of the metrics are trained with local neighboring data pairs. Similar to data association, we find that temporal windows can effectively find the corresponding local neighborhood intra-camera or inter-camera. Hence, we use temporal windows for sampling training data pairs in the proposed LAAM.

\textbf{Intra-camera metric.} 
For data associations in SCT, we train an intra-camera metric to provide similarity estimation between tracklets. The metric network takes tracklet features as input for both training and testing. 
During training, the tracklet features are computed on ground truth images, while during testing,  the tracklet features are computed on pedestrian detections.

In training, we sample local neighboring data pairs within a small temporal duration of $\tau_S$ from the target camera.
As shown in Fig.~\ref{fig:TLML_sampling}, for every tracklet (yellow box) and the corresponding  feature $\bm{f}_i$,
we first randomly select data pair being of the same identity or not. These same identity data pairs and different identity data pairs are denoted as positive pairs and negative pairs, respectively. 
Note that the positive/negative pairs are generated with a $1:1$ ratio for data balance. We then choose the corresponding tracklet feature $\bm{f}_j$ for the positive/negative pair. 
For positive pairs, we sample a tracklet (green box) that belongs to the same identity $n$ as the first tracklet, within the $\tau_S$-sized window. 
For negative pairs, we sample a tracklet (red box) that belongs to a different identity, within the same data sampling window. 
Either way, we end up with a tracklet feature pair $\bm{f}_i$ and $\bm{f}_j$.
At last, we feed the absolute difference vector $\bm{f}=\left|\bm{f}_i - \bm{f}_j\right|$ into the metric network as input. 



\begin{figure}[t]
    \centering
    \begin{subfigure}[b]{0.23\textwidth}
        \includegraphics[width=\textwidth]{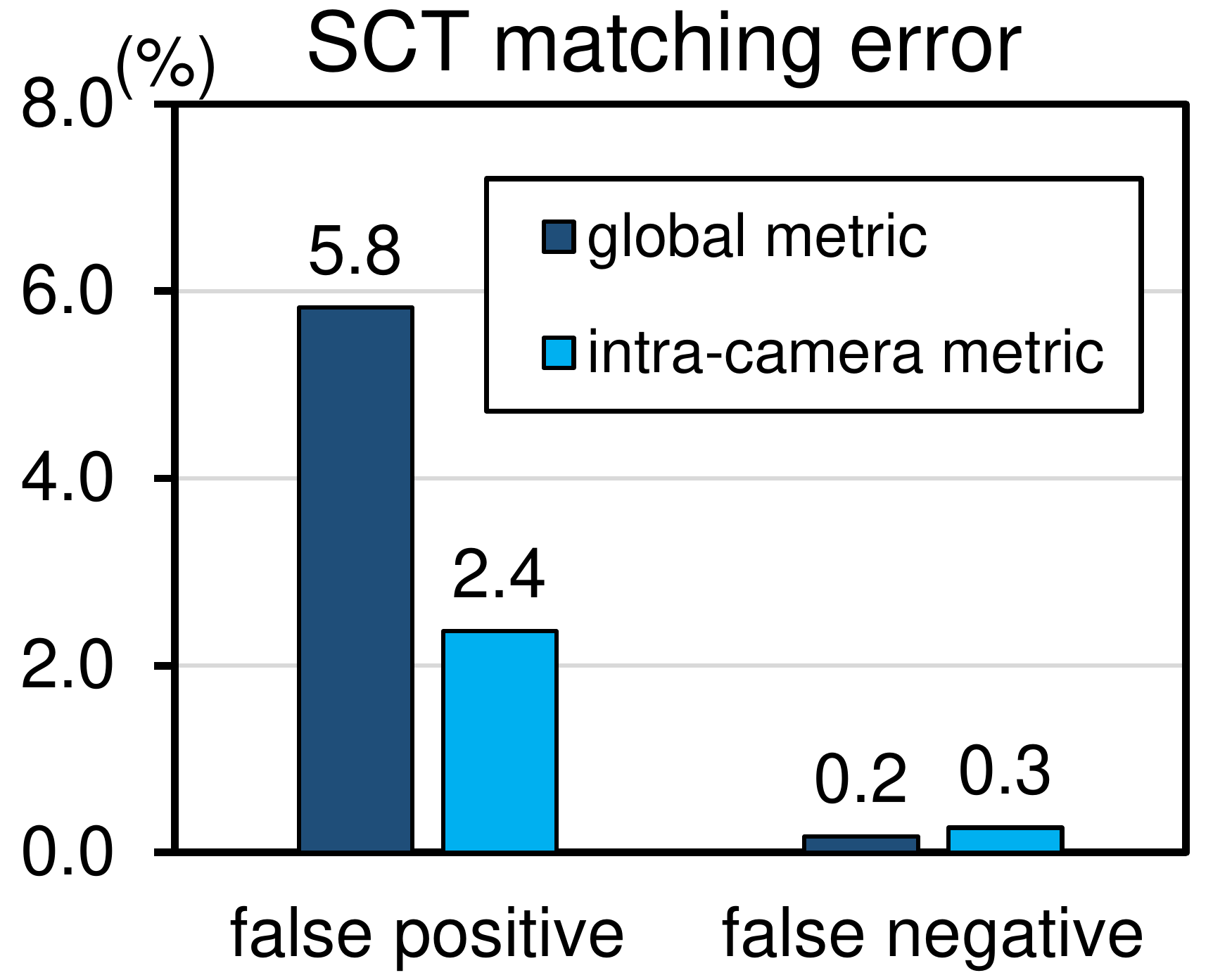}
    \end{subfigure}
    ~ 
    \begin{subfigure}[b]{0.23\textwidth}
        \includegraphics[width=\textwidth]{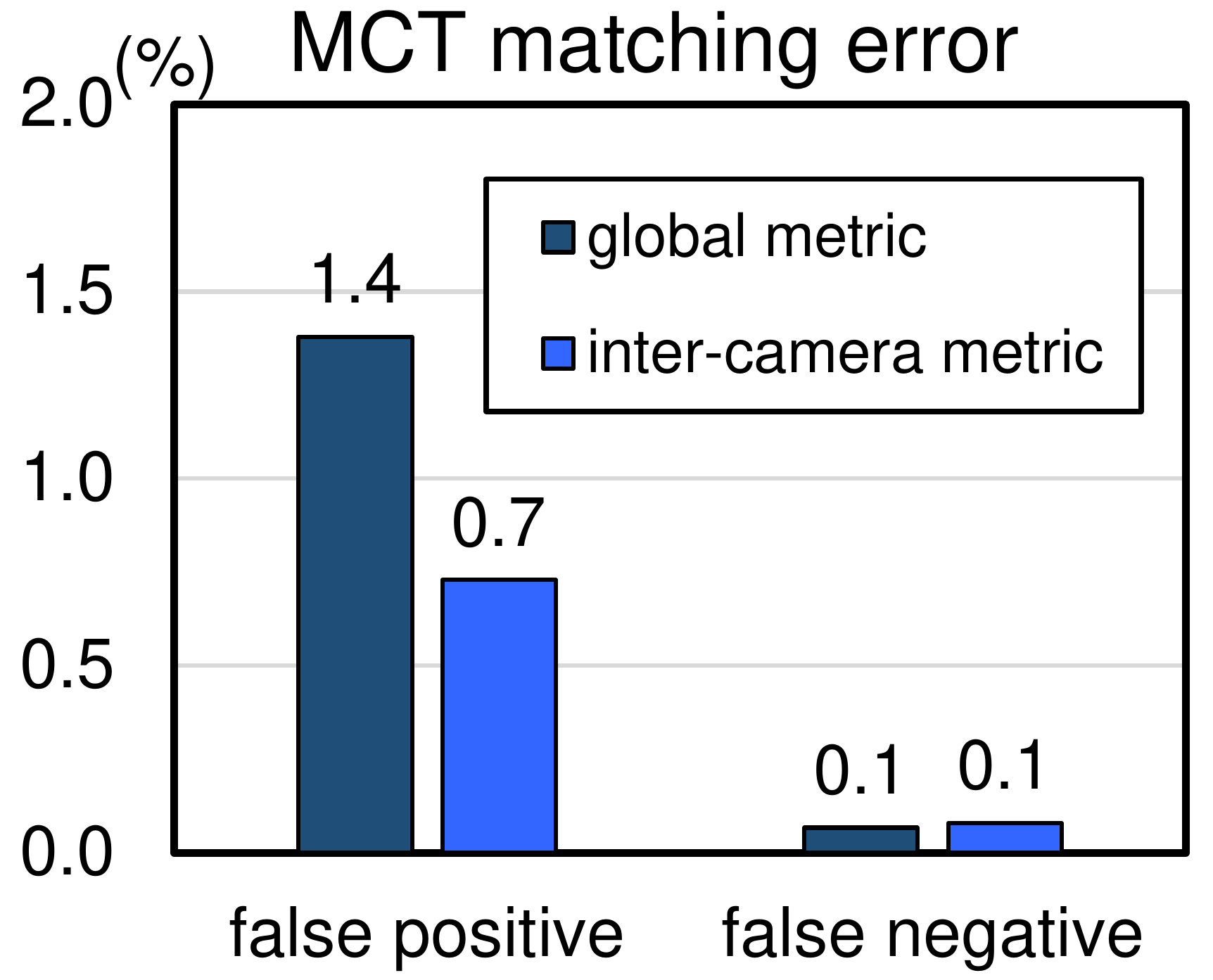}
    \end{subfigure}
\vspace{-2mm}
\caption{Matching error comparison between the global metric and the proposed LAAM. We report false positive rate and false negative rate on the validation set.}
\vspace{-3mm}
\label{fig:fpfn}
\end{figure}

\textbf{Inter-camera metric.}
For data association in MCT, we train an inter-camera metric to provide similarity estimation between single camera trajectories. The metric network takes tracklet/trajectory features as input for training and testing, respectively. 
During training, we use tracklet features instead of trajectory features, due to the scarcity of single camera trajectories. Same as intra-camera metric, tracklet features for training are computed on ground truth images. 
During testing, trajectory features are computed on pedestrian detections.

The construction of cross-camera data pairs is very similar to that of within-camera data pairs, except for the following differences. 
First, the data pair is chosen within a $\tau_M$-frame-long data sampling window across cameras. This cross-camera sampling window length $\tau_M$ is usually bigger than the within-camera sampling window length $\tau_S$. 
Second, the positive/negative data pair sampling mechanism is different. 
For positive data pairs, we sample a tracklet (green box) that belongs to the same identity $n$ as the first tracklet within the $\tau_M$-sized window, and we require the positive tracklet to be sampled from a different camera. 
For negative data pairs, we sample a tracklet (red box) that belongs to different identities, from a random camera within the temporal sampling window $\tau_M$.

\subsection{Discussion}\label{sec:discuss}
\textbf{Comparison between data sampling windows and temporal sliding windows.} 
Data sampling windows and temporal sliding windows share some similarities. Both of them are used to restrict the data pool to its local neighborhood. 
However, there are a major difference. Temporal sliding windows are used in data association. In contrast, data sampling windows are used in data sampling to train the locality aware appearance metric (LAAM).  

\textbf{Data sampling window lengths.} The lengths of the data sampling window in LAAM are different for intra-camera metric and inter-camera metric. The window length $\tau_S$ for within-camera data sampling is similar to the average trajectory duration. On the other hand, window length $\tau_M$ for cross-camera data sampling should be long enough to cover local neighboring trajectories of the same identity in different cameras. The influence of sampling window length on LAAM training are shown in Fig.~\ref{fig:perf}.

\textbf{Comparison between global metric and LAAM.}
We validate LAAM through statistics comparisons. The matching errors during SCT and MCT are shown in Fig.~\ref{fig:fpfn}. We use the intra-camera metric for SCT and inter-camera metric for MCT. Using the proposed method, the false positive rate is significantly lower than the global metric, while the false negative rate remains very similar. 

\textbf{Extreme cases.} First, when the video frame rate is extremely low, unless they are returning, each target will only appear in one camera once. Under the circumstances, SCT will be completely removed, and thus the intra-camera metric will be obsolete. However, since the trajectory continuity still holds and the topology does not change, the locality in MCT will not be influenced. Thus, the inter-camera metric is still useful. 
Second, when the scenario is in open topology, targets travel to all cameras at the same probability. This time, the inter-camera metric will fall back to the global metric. However, the SCT data association is still local, and thus the intra-camera metric is useful.

\section{Experiment} \label{results}

\subsection{Dataset and Evaluation Protocol}
\textbf{Dataset.} This paper uses the DukeMTMC dataset~\cite{ristani2016performance} and CityFlow dataset~\cite{tang2019cityflow} to evaluate the proposed metric. 
DukeMTMC is a pedestrian tracking dataset. It contains 1080p 60fps videos from 8 cameras on a school campus. 
CityFlow is a vehicle tracking dataset. It has a low frame rate (10fps), severe occlusion, and fast-moving vehicles from 40 cameras, spanning over 2km.
For simplicity, if not mentioned, we refer to the ``validation set'', ``test (easy)'', and ``test (hard)'' as that of the DukeMTMC dataset.

We also use the DukeMTMC-reID \cite{zheng2017unlabeled} and Market-1501 \cite{zheng2015scalable} datasets to evaluate re-ID appearance features. 

\textbf{Training and validation sets for DukeMTMC.}
In the DukeMTMC experiment, we use the first 40 minutes of the ground truth as the train set and the remaining 10 minutes as the validation set. 
For both the validation and online testing, we only use the train on the train set.

\textbf{Evaluation protocol.} 
For MTMCT, following \cite{ristani2016performance}, we use IDF1, IDP, and IDR as evaluation metrics. Note that CityFlow only evaluates MCT. 
For both DukeMTMC and CityFlow datasets, we evaluate on their online test set. 
For re-ID evaluation, we adopt the rank-1 accuracy and mean average precision (mAP)~\cite{zheng2015scalable} evaluation protocol.

\begin{table}[t]
    \centering
\resizebox{\linewidth}{!}{
\begin{tabular}{l||l|l}
\toprule
Method/variant               & SCT similarity      &  MCT similarity   \\                                                                                                                                                                                                                 \hline
``Baseline''            & Euclidean distance based & Euclidean distance based\\                                                                                                                                        \hline
``Global metric''       & Global metric     &   Global metric \\
\hline
``intra/global''        & Intra-camera metric & Global metric\\                                                                                                                                                                             \hline
``global/inter''        & Global metric  & Inter-camera metric  \\                                                                                                                                                                          \hline
``intra/intra''         & Intra-camera metric &      Intra-camera metric\\                                                                                \hline
``inter/inter''         & Inter-camera metric & Inter-camera metric \\                                                                                     \hline
`inter/intra''         & Inter-camera metric & Intra-camera metric \\                                                                                     \hline
``intra/inter'' (ours)         & Intra-camera metric & Inter-camera met   \\ \bottomrule
\end{tabular}
}
\vspace{-1mm}
\caption{Methods/variants compared in our experiment.}
\vspace{-3mm}
\label{tab:alias}
\end{table}

\subsection{Implementation Details}\label{implement}

\textbf{Re-ID features.} 
On DukeMTMC, we use three globally learned re-ID features, namely, the ID-discriminative embedding (IDE)~\cite{zheng2015scalable}, the triplet feature~\cite{hermans2017defense}, and the part-based convolutional baseline (PCB)~\cite{sun2018PCB}. 
To train the three networks, we use the following settings. The input image is resized to $384 \times 128$.
Random erasing~\cite{zhong2017random} is employed for data augmentation. We use $\frac{1}{60}$ of the ground truth images (1 frame every second) as training data for faster convergence. In fact, training re-ID features with fewer frames 1) enables fast convergence and 2) does not lead to accuracy drop. 
We use ResNet-50 \cite{he2016deep} pre-trained on ImageNet~\cite{deng2009imagenet} as the backbone for the three models. 

On CityFlow, we use a DenseNet-121~\cite{huang2017densely} based re-ID feature with softmax and triplet loss. We train on the provided vehicle re-ID dataset for CityFlow. 

\textbf{Baseline MTMCT tracker.}
Our baseline is developed from DeepCC \cite{ristani2018features}, with the one modification. 
We allow a target returning to the same camera. This also helps recognize a target after long-time occlusion, which is difficult for SCT with a short temporal sliding window. 
On DukeMTMC, each tracklet has $40$ frames. The temporal sliding window lengths for SCT and MCT are $150$ frames and $6,000$ frames, respectively. 
On CityFlow, we set the tracklet length to 10 frames. Temporal sliding windows for SCT and MCT are set to 500-frame-long and 2400-frame-long, respectively. 
$\mu_p$ and $\mu_n$ are calculated from the train set in both datasets.

\textbf{Metric learning settings.} The proposed locality aware appearance metric is trained with tracklet features, average pooled from ground truth image re-ID features. The learning rate is set to $1\times10^{-4}$ for the first 30 epochs, and then decays to $0.1\times$ for the last 10 epochs. The batch size is set to $64$. We use the cross-entropy loss to train the metric network. 
On DukeMTMC, within-camera data sampling window length $\tau_S$ is $600$ frames, whereas cross-camera data sampling window length $\tau_M$ is $2,400$ frames. 
On CityFlow, we set $\tau_S=30$ and $\tau_M=500$.

\textbf{Method variants and notations.} In Table~\ref{tab:alias}, we present some descriptions and notations of the methods to be evaluated in the experiment. 
The baseline uses the Euclidean distance for similarity estimation (Eq.~\ref{eq:w_a}). 
Similarity estimation in all the other variants is calculated as Eq.~\ref{eq:w_metric}.  
The global metric also adopts the structure in Fig.~\ref{fig:TLML_sturcture}. 
``intra/inter'' is the proposed full system.

\subsection{Evaluation of Re-ID Features}
The performance of re-ID features used in existing MTMCT works and in our paper is summarized in Table~\ref{tab:reid}. First, we find the accuracy of IDE is on par with the triplet feature and is lower than PCB. It is consistent with the observation in \cite{sun2018PCB}. Second, comparing with the re-ID descriptors used in previous works, our feature extractors are competitive on both the DukeMTMC-reID \cite{zheng2017unlabeled} and Market-1501 \cite{zheng2015scalable} datasets. For example, on Market-1501, the rank-1 accuracy is 92.0\% for PCB, which is consistent with the reports in \cite{sun2018PCB} and is close to the accuracy in \cite{zhang2017multi}. 
We further assess their influence on LAAM in Table~\ref{tab:val}. In the following experiment, if not specified, we use IDE as the default pedestrian descriptor due to its good accuracy and easy implementation.

\begin{table}[t]
    \centering
\resizebox{\linewidth}{!}{
\begin{tabular}{l|cc|cc}
\toprule
\multirow{2}{*}{Features} & \multicolumn{2}{c|}{DukeMTMC-reID}                        & \multicolumn{2}{c}{Market-1501}                       \\ \cline{2-5} 
                             & \multicolumn{1}{c}{rank-1} & \multicolumn{1}{c|}{mAP} & \multicolumn{1}{c}{rank-1} & \multicolumn{1}{c}{mAP} \\ \hline
DeepCC~\cite{ristani2018features}     & 79.8             & 63.4            & 89.5            & 75.7          \\ \hline
MTMC\_ReID~\cite{zhang2017multi} & 81.9           & N/A             & \textbf{93.9 }           & N/A           \\ \hline
TAREIDMTMC~\cite{jiang2018online} & 81.6             & \textbf{72.3}             & 87.2           & 76.4        \\ \hline\hline
{IDE}        & 79.7             & 62.9            & 87.6           & 72.2           \\ \hline
Triplet             & 81.3              & 66.4            & 89.3            & 76.3           \\ \hline
PCB         & \textbf{82.9}           & 68.6           & 92.0           & \textbf{78.2}           \\\bottomrule
\end{tabular}
}
\vspace{-1mm}
\caption{Rank-1 accuracy (\%) and mAP (\%) of re-ID features on the Market-1501 and DukeMTMC-reID datasets. The three features (IDE, Triplet and PCB) we use in this paper have competitive accuracy in re-ID. }
\vspace{-3mm}
\label{tab:reid}
\end{table}

\subsection{Evaluation of the Proposed Method} 
In this section, we summarize the results obtained by the proposed LAAM and compare it with its variants and the state-of-the-art methods. 

\textbf{Improvement over the baseline tracker and global metric learning.}
We first compare our method against baseline and global metric. Results are shown in Table~\ref{tab:val}, Table~\ref{tab:cityflow}, and Table Table~\ref{tab:test}. We have two observations. 

First, the global metric learning does not improve over the baseline. For example, on the validation set, compared with the Euclidean distance based baseline, applying global metric on IDE feature changes the IDF1 by -0.5\% in SCT and by +0.2\% in MCT. Under the same setting, on the easy test set, IDF1 accuracy of the global metric is equal to the baseline in SCT and is +0.3\% higher in MCT. On the CityFlow dataset, global metric improves the MCT IDF1 by +0.5\%.
These results indicate that global metric learning does not bring significant benefits. This is because both the baseline and global metric are trained on the global train set, so their discriminative abilities are very close. 

Second, the full LAAM method brings a consistent and non-trivial improvement over the baseline and global metric. On the validation set, for example, the full method ``LAAM (intra/inter)'' has a +2.4\% IDF1 improvement over the baseline on the MCT task using IDE as the feature. \textbf{On test (hard), our method using IDE excels the baseline by +6.9\% in terms of IDF1, a significant improvement. On CityFlow, the proposed method improves MCT IDF1 by +6.4\%. }
It demonstrates the effectiveness and generalization ability of our full method in terms of its ability in improving baseline accuracy, thus validating the proposed metric to some extent. 

\textbf{Impact of different re-ID features.}
Tracking accuracy based on different re-ID features is summarized in Table \ref{tab:val}. Under both the SCT and MCT task, we find that the tracking performance of IDE, the triplet feature, and PCB is similar. This finding is consistent with a previous report \cite{ristani2018features}: improvement in re-ID accuracy can have a diminishing improvement on the MTMCT system. 
The main reason is that the appearance variation in tracking is much smaller than that in re-ID. For example, in MCT, the gallery in a temporal sliding window might have dozens of images, while that in re-ID has over 10k images. With a much smaller gallery, there is less requirement on feature's discriminative ability, and PCB would have a similar matching accuracy with IDE. Moreover, MTMCT also has several other components besides feature-based matching. Imperfectness in these components reduces the improvement brought about by the re-ID features. 


\begin{table}[t]
\centering
\resizebox{\linewidth}{!}{
\begin{tabular}{l|ll|ll|ll}
\toprule
\multirow{3}{*}{Variant} & \multicolumn{6}{c}{Validation set IDF1 results}                                                                                                                         \\ \cline{2-7} 
                         & \multicolumn{2}{c|}{{IDE}}                         & \multicolumn{2}{c|}{triplet}                         & \multicolumn{2}{c}{PCB}                        \\ \cline{2-7} 
                         & \multicolumn{1}{c}{SCT} & \multicolumn{1}{c|}{MCT} & \multicolumn{1}{c}{SCT} & \multicolumn{1}{c|}{MCT} & \multicolumn{1}{c}{SCT} & \multicolumn{1}{c}{MCT} \\ \hline
Baseline                 & 86.4                    & 81.4                     & 86.2                    & 80.9                     & 85.8                    & 80.6                    \\ \hline
Global metric                & 85.9                    & 81.6                     & 84.1                    & 79.7                     & 87.4                    & 81.6                    \\ \hline
LAAM (intra/global)                    & 87.8                    & 83.1                     & 87.6                    & 83.9                     &  87.1                   & 82.4                    \\ \hline
LAAM (global/inter)                    & 86.0                  & 81.6                 & 84.5                   & 79.9                     & 87.9                    & 82.5                    \\ \hline
LAAM (intra/intra)                    & 87.8                    & 83.4                     & 87.8                    & 84.2                     & 87.7                    & 82.4                    \\ \hline
LAAM (inter/inter)                    & 86.9                    & 82.5                     & 87.4                    & 83.9                     & 87.5                    & 82.5                    \\ \hline
LAAM (inter/intra)                    & 86.3                    & 81.6                     & 85.6                    & 82.1                     & 87.5                    & 82.1                    \\ \hline
\textbf{LAAM (intra/inter)}              & \textbf{87.9}           & \textbf{83.8}            & \textbf{87.9}           & \textbf{84.5}            & \textbf{87.7}           & \textbf{82.9}           \\ \bottomrule
\end{tabular}
}
\vspace{-1mm}
\caption{IDF1 accuracy on the DukeMTMC validation set. Three re-ID features are evaluated under various methods.}
\vspace{-3mm}
\label{tab:val}
\end{table}

\textbf{Comparison with variants and ablation study.} 
We replace the intra-camera metric with the global metric or the inter-camera metric; we also replace the inter-camera metric with the global metric or the intra-camera metric. Results are shown in Table~\ref{tab:val} and Table~\ref{tab:test}.

First, we show both metrics are necessary. In Table \ref{tab:val}, when replacing intra-camera metric with the global metric, IDF1 based on the IDE feature drops by 1.9\% and 2.2\% on SCT and MCT, respectively. A similar but smaller accuracy drop can be observed when the inter-camera metric is replaced with the global metric. The drop is consistent when using different re-ID features. These results show that both the intra-camera and inter-camera metrics are necessary components in our system.

Second, from the ablation studies, the removal of the intra-camera metric causes a larger accuracy drop. The possible reason might be the variance gap between local data association in tracking and global matching in re-ID is bigger in SCT and smaller in MCT. Within a single camera, appearance variance of a target is very small. Between neighboring camera pairs, the appearance has a larger variance (still smaller than global). In a global sense, the appearance changes are the largest. 
Since there is a largest gap between SCT (local) and re-ID (global) matching, the intra-camera metric has a larger improvement.

\begin{table}[t]\small
\centering
\setlength{\tabcolsep}{5mm}{
\begin{tabular}{l|l|l|l}
\toprule
\multirow{2}{*}{Variant} & \multicolumn{3}{c}{CityFlow test set MCT results}                                                       \\ \cline{2-4} 
                                  & \multicolumn{1}{c|}{IDF1} & \multicolumn{1}{c|}{IDP} & \multicolumn{1}{c}{IDR} \\ \hline
Baseline                          & 56.6                      & 53.3                     & 60.7                    \\ \hline
Global metric                     & 57.1                      & 54.4                     & 60.7                    \\ \hline
\textbf{LAAM (intra/inter) }               & \textbf{63.0}             & \textbf{60.7}            & \textbf{66.0}           \\ \bottomrule
\end{tabular}
}
\vspace{-1mm}
\caption{CityFlow online test set results. Note that CityFlow dataset only evaluate multi-camera tracking. The proposed method yields substantial accuracy increase.}
\vspace{-3mm}
\label{tab:cityflow}
\end{table}

\begin{table*}[t]
\centering
\resizebox{\linewidth}{!}{
\small
\setlength{\tabcolsep}{6.8pt} 
\begin{tabular}{l|l|lll|lll|lll|lll} 
\toprule
\multirow{3}{*}{Tracker}     & \multirow{3}{*}{Detector} & \multicolumn{6}{c|}{test (easy)}                                                                                                                          & \multicolumn{6}{c}{test (hard)}                                                                                                                           \\ 
\cline{3-14}
                             &                           & \multicolumn{3}{c|}{SCT}                                                      & \multicolumn{3}{c|}{MCT}                                                      & \multicolumn{3}{c|}{SCT}                                                      & \multicolumn{3}{c}{MCT}                                                       \\ 
\cline{3-14}
                             &                           & \multicolumn{1}{c}{IDF1} & \multicolumn{1}{c}{IDP} & \multicolumn{1}{c|}{IDR} & \multicolumn{1}{c}{IDF1} & \multicolumn{1}{c}{IDP} & \multicolumn{1}{c|}{IDR} & \multicolumn{1}{c}{IDF1} & \multicolumn{1}{c}{IDP} & \multicolumn{1}{c|}{IDR} & \multicolumn{1}{c}{IDF1} & \multicolumn{1}{c}{IDP} & \multicolumn{1}{c}{IDR}  \\ 
\hline
BIPCC$^*$~\cite{ristani2016performance}                        & DPM~ \cite{felzenszwalb2010object}                      & 70.1                     & 83.6                    & 60.4                     & 56.2                     & 67.0                    & 48.4                     & 64.5                     & 81.2                    & 53.5                     & 47.3                     & 59.6                    & 39.2                     \\ 
\hline
MTMC\_CDSC~\cite{tesfaye2017multi}                    & DPM                       & 77.0                     & 87.6                    & 68.6                     & 60.0                     & 68.3                    & 53.5                     & 65.5                     & 81.4                    & 54.7                     & 50.9                     & 63.2                    & 42.6                     \\ 
\hline
MYTRACKER$^*$~\cite{yoon2018multiple}                    & DPM                       & 80.3                     & 87.3                    & 74.4                     & 65.4                     & 71.1                    & 60.6                     & 63.5                     & 73.9                    & 55.6                     & 50.1                     & 58.3                    & 43.9                     \\ 
\hline
MTMC\_ReIDp~\cite{zhang2017multi}                  & DPM                       & 79.2                     & 89.9                    & 70.7                     & 74.4                     & 84.4                    & 66.4                     & 71.6                     & 85.3                    & 61.7                     & 65.6                     & 78.1                    & 56.5                     \\ 
\hline
TAREIDMTMC$^*$~\cite{jiang2018online}                   & Mask R-CNN~\cite{he2017mask}                & 83.8                     & 87.6                    & 80.4                     & 68.8                     & 71.8                    & 66.0                     & 77.9                     & 86.6                    & 70.7                     & 61.2                     & 68.0                    & 55.5                     \\ 
\hline
DeepCC~\cite{ristani2018features}                       & OpenPose~\cite{cao2018openpose}                 & 89.2                     & 91.7                    & 86.7                     & 82.0                     & 84.4                    & 79.8                     & 79.0                     & 87.4                    & 72.0                     & 68.5                     & 75.9                    & 62.4                     \\ 
\hline
MTMC\_ReID~\cite{zhang2017multi}                   & Faster R-CNN~\cite{ren2015faster}             & 89.8                     & 92.0                    & 87.7                     & 83.2                     & 85.2                    & 81.2                     & 81.2                     & 89.4                    & 74.5                     & 74.0                     & 81.4                    & 67.8                     \\ 
\hline\hline
Baseline                     & \multirow{3}{*}{OpenPose} & 91.3                     & 91.8                    & 90.9                     & 87.4                     & 87.8                    & 87.0                     & 83.7                     & 88.8                    & 79.1                     & 75.4                     & 80.0                    & 71.3                     \\ 
\cline{1-1}\cline{3-14}
Global metric                &                           & 91.3                     & 92.2                    & 90.4                     & 87.7                     & 88.6                    & 86.8                     & 82.7                     & 89.2                    & 77.1                     & 76.2                     & 82.2                    & 71.0                     \\ 
\cline{1-1}\cline{3-14}
\cline{1-1}\cline{3-14}
\cline{1-1}\cline{3-14}
\textbf{LAAM (intra/inter)}  &                           & \textbf{92.5}            & \textbf{93.0}           & \textbf{92.0}            & \textbf{88.6}            & \textbf{89.0}           & \textbf{88.1}            & \textbf{85.8}                     & \textbf{91.1}                    & \textbf{81.1}            & \textbf{82.3}            & \textbf{87.4}           & \textbf{77.8}            \\
\bottomrule
\end{tabular}
}
\vspace{-2mm}
\caption{DukeMTMC online test set results. Methods with $^*$ are online tracking methods. ``LAAM (intra/inter)'' refers to the proposed method. On both the easy and hard test sets, our method yields very competitive accuracy.}
\vspace{-3mm}
  \label{tab:test}
\end{table*}

Third, we show that the two metrics are not interchangeable. In Table~\ref{tab:val}, when we replace the intra-camera metric with the inter-camera metric, as in ``LAAM (inter/inter)'', IDF1s drop by 1.0 and 1.3\% on the validation set SCT and MCT. Reversely, when we compare the full method with ``LAAM (intra/intra)'', SCT and MCT IDF1s drop by 0.1\% and 0.4\%, respectively. When we swap intra-camera metric and inter-camera metric, IDF1s drop by 1.6\% on SCT and 2.2\% on MCT. These drops are consistent with different re-ID features. This suggests that the two metrics work best on SCT and MCT, respectively. This is consistent with their respective data sampling methods (Section~\ref{sec:two_metrics}). 


\textbf{Comparison with the state-of-the-art methods.}
In Table \ref{tab:test}, we compare our method using the IDE features with state-of-the-art methods. We have two observations.
First, baseline tracker is very competitive. The baseline itself surpasses many existing methods like \cite{ristani2018features,zhang2017multi}. This demonstrates the effectiveness of our modified tracker and re-ID feature. 
Second, LAAM further improves over the competitive baseline tracker and achieves new state-of-the-art accuracy on both the easy and hard test sets. On the easy test set, \textbf{we obtain 92.5\% and 88.6\% in IDF1 on SCT and MCT, respectively.} These numbers are +2.7\% and 5.4\% higher than the second-best results \cite{zhang2017multi}. On the hard test set, \textbf{our IDF1 scores are 85.8\% and 82.3\% on SCT and MCT, respectively.} This is +4.6\% and +8.3\% higher than the second-best method \cite{zhang2017multi}. These comparisons indicate that our method is particularly advantageous in MCT and the challenging scenarios.

\begin{figure}[t]
    \centering
    \begin{subfigure}[b]{0.23\textwidth}
        \includegraphics[width=\textwidth]{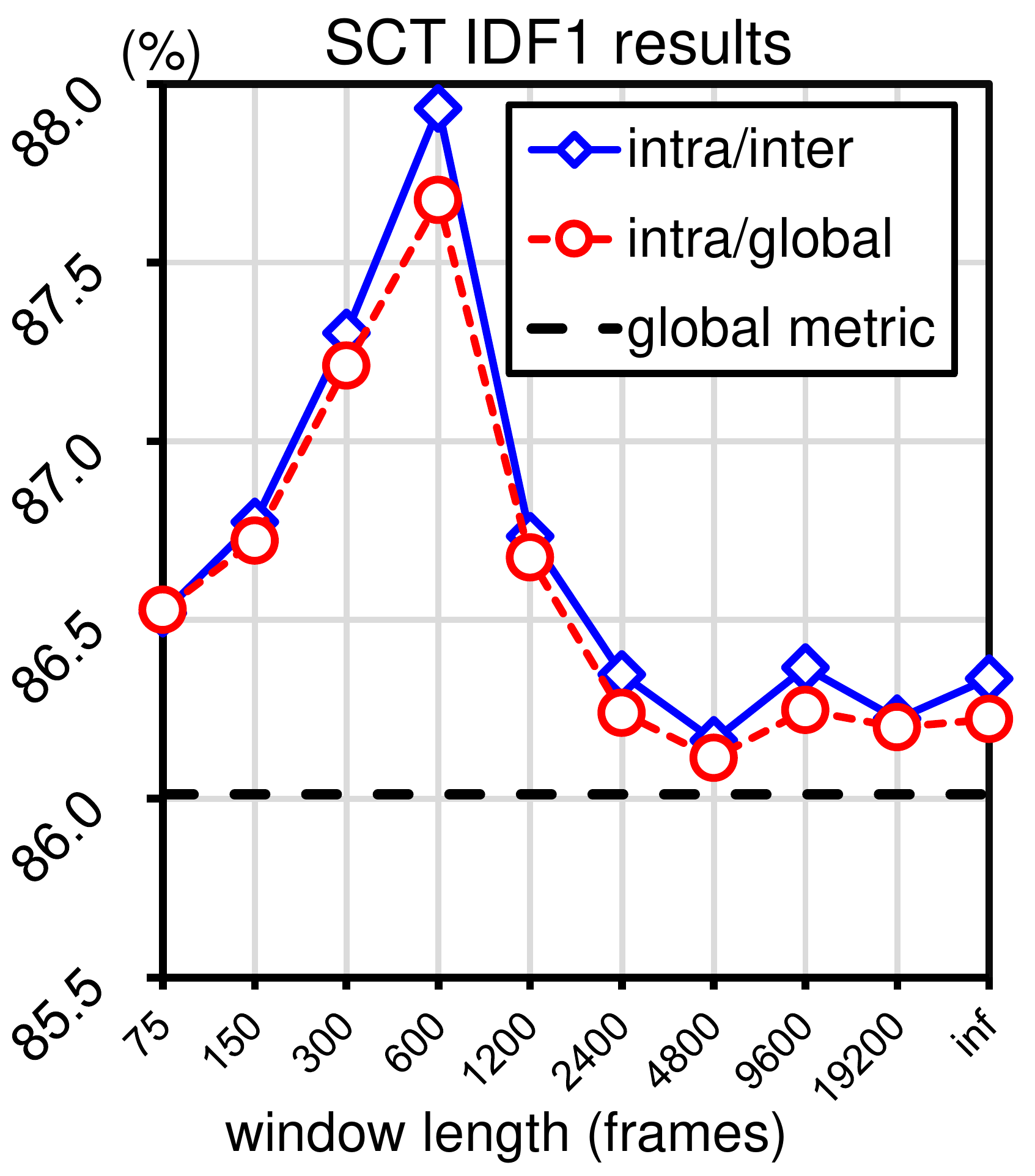}
        \label{fig:sct perf}
    \end{subfigure}
    \hfill 
    \begin{subfigure}[b]{0.23\textwidth}
        \includegraphics[width=\textwidth]{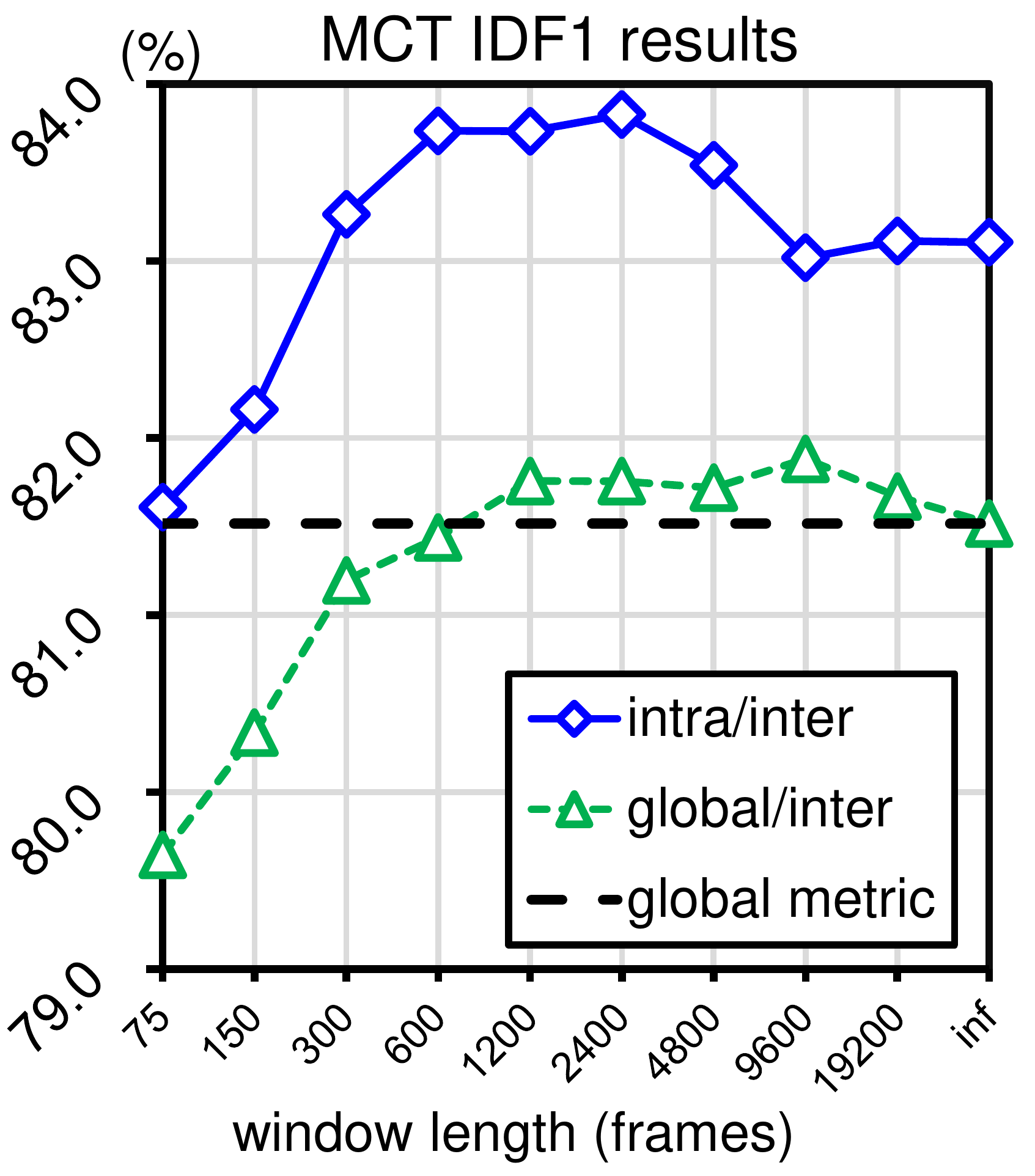}
        \label{fig:mct perf}
    \end{subfigure}
\vspace{-3mm}
\caption{Influence of data sampling window length of LAAM on SCT and MCT. The dashed line is the accuracy of the global metric. IDF1 on the validation set is reported.}
\vspace{-3mm}
\label{fig:perf}
\end{figure}

\textbf{Parameter analysis.} 
We assess the impact of data sampling window lengths in Fig.~\ref{fig:perf} as key parameter analysis. 
A short sampling window may significantly reduce the choices of training pairs, leaving the metric more prone to overfitting.
On the other hand, a long sampling window no longer underpins locality. 
From the results, the best within-camera and cross-camera sampling window sizes are $600$ and $2,400$, respectively. 

We observe two other phenomena worth noticing. First, from Fig. \ref{fig:perf}, the inter-camera metric designed for MCT improves SCT as well. In fact, our tracker allows returning targets, so correctly labeling of these returning targets improves SCT accuracy. Second, from Fig. \ref{fig:perf}, the inter-camera metric is inferior to the global metric under short windows. This is because when the cross-camera sampling window is shorter than the camera transition time, there will not be sufficient cross-camera training samples. 

\textbf{Computation complexity.} 
The metric network takes 20 minutes to train on a server with a GTX 1080ti GPU. During testing, the CNN features are extracted with GPU, and the tracker including the metric similarity score is computed on the 3.2Ghz Intel Xeon CPU. In fact, frequently calling GPU for the 3-layer metric network takes more time. In testing, creating tracklets takes 912 seconds. Computing single camera trajectories take 447 seconds and 520 seconds in the baseline and LAAM, respectively. Computing cross-camera tracks take 105 seconds in both the baseline and our method. Overall, baseline spends 1,464 seconds in testing, whereas LAAM spends 1,537 seconds. Our method causes 5\% more testing time, which is acceptable.

In MTMCT, better similarity estimation usually makes data association easier. Therefore, although LAAM consumes more time in similarity estimation, it saves time in data association by providing more accurate similarity scores. In SCT, data association is relatively easy, and most of the time is spent in similarity computation. In MCT, data association is more difficult and dominates the computation time. As a result, compared with the baseline, our method is slower in SCT and spends a similar time on MCT. 

\section{Conclusion}
This paper draws novel insights towards the inherent differences between re-ID and MTMCT. That is, re-ID is a global matching problem, while MTMCT is based on local matching. This difference compromises the effectiveness of directly applying global re-ID appearance features in local matching of MTMCT. 
This paper investigates how to effectively fit \textit{global} appearance features into \textit{local} matching in tracking. To this end, we propose the locality aware appearance metric (LAAM), which uses a novel training data sampling strategy. Given globally learned re-ID features, pairs of training data are sampled from their local neighborhood. For single camera tracking (SCT), local neighborhood refers to the consecutive frames within a single camera; for multi-camera tracking (MCT), it refers to neighboring cameras that a target may appear successively. 
On two MTMCT datasets, we show that LAAM leads to significant improvements over the baseline, and report new state-of-the-art tracking accuracy on DukeMTMC.



{\small
\bibliographystyle{ieee_fullname}
\bibliography{egbib}

\begin{thebibliography}{10}\itemsep=-1pt

\bibitem{barbosa2018looking}
Igor~Barros Barbosa, Marco Cristani, Barbara Caputo, Aleksander Rognhaugen, and
  Theoharis Theoharis.
\newblock Looking beyond appearances: Synthetic training data for deep cnns in
  re-identification.
\newblock {\em Computer Vision and Image Understanding}, 167:50--62, 2018.

\bibitem{bellet2013survey}
Aur{\'e}lien Bellet, Amaury Habrard, and Marc Sebban.
\newblock A survey on metric learning for feature vectors and structured data.
\newblock {\em arXiv preprint arXiv:1306.6709}, 2013.

\bibitem{berclaz2011multiple}
Jerome Berclaz, Francois Fleuret, Engin Turetken, and Pascal Fua.
\newblock Multiple object tracking using k-shortest paths optimization.
\newblock {\em IEEE transactions on pattern analysis and machine intelligence},
  33(9):1806--1819, 2011.

\bibitem{brendel2011multiobject}
William Brendel, Mohamed Amer, and Sinisa Todorovic.
\newblock Multiobject tracking as maximum weight independent set.
\newblock In {\em Computer Vision and Pattern Recognition (CVPR), 2011 IEEE
  Conference on}, pages 1273--1280. IEEE, 2011.

\bibitem{cai2014exploring}
Yinghao Cai and Gerard Medioni.
\newblock Exploring context information for inter-camera multiple target
  tracking.
\newblock In {\em Applications of Computer Vision (WACV), 2014 IEEE Winter
  Conference on}, pages 761--768. IEEE, 2014.

\bibitem{cao2018openpose}
Zhe Cao, Gines Hidalgo, Tomas Simon, Shih-En Wei, and Yaser Sheikh.
\newblock Open{P}ose: realtime multi-person 2{D} pose estimation using {P}art
  {A}ffinity {F}ields.
\newblock In {\em arXiv preprint arXiv:1812.08008}, 2018.

\bibitem{chari2015pairwise}
Visesh Chari, Simon Lacoste-Julien, Ivan Laptev, and Josef Sivic.
\newblock On pairwise costs for network flow multi-object tracking.
\newblock In {\em CVPR}, volume~20, page~15, 2015.

\bibitem{cheng2016person}
De Cheng, Yihong Gong, Sanping Zhou, Jinjun Wang, and Nanning Zheng.
\newblock Person re-identification by multi-channel parts-based cnn with
  improved triplet loss function.
\newblock In {\em Proceedings of the IEEE Conference on Computer Vision and
  Pattern Recognition}, pages 1335--1344, 2016.

\bibitem{choi2015near}
Wongun Choi.
\newblock Near-online multi-target tracking with aggregated local flow
  descriptor.
\newblock In {\em Proceedings of the IEEE international conference on computer
  vision}, pages 3029--3037, 2015.

\bibitem{collins2012multitarget}
Robert~T Collins.
\newblock Multitarget data association with higher-order motion models.
\newblock In {\em Computer Vision and Pattern Recognition (CVPR), 2012 IEEE
  Conference on}, pages 1744--1751. IEEE, 2012.

\bibitem{das2014consistent}
Abir Das, Anirban Chakraborty, and Amit~K Roy-Chowdhury.
\newblock Consistent re-identification in a camera network.
\newblock In {\em European Conference on Computer Vision}, pages 330--345.
  Springer, 2014.

\bibitem{dehghan2015gmmcp}
Afshin Dehghan, Shayan Modiri~Assari, and Mubarak Shah.
\newblock Gmmcp tracker: Globally optimal generalized maximum multi clique
  problem for multiple object tracking.
\newblock In {\em Proceedings of the IEEE Conference on Computer Vision and
  Pattern Recognition}, pages 4091--4099, 2015.

\bibitem{dehghan2015target}
Afshin Dehghan, Yicong Tian, Philip~HS Torr, and Mubarak Shah.
\newblock Target identity-aware network flow for online multiple target
  tracking.
\newblock In {\em Proceedings of the IEEE Conference on Computer Vision and
  Pattern Recognition}, pages 1146--1154, 2015.

\bibitem{deng2009imagenet}
Jia Deng, Wei Dong, Richard Socher, Li-Jia Li, Kai Li, and Li Fei-Fei.
\newblock Imagenet: A large-scale hierarchical image database.
\newblock In {\em Computer Vision and Pattern Recognition, 2009. CVPR 2009.
  IEEE Conference on}, pages 248--255. Ieee, 2009.

\bibitem{fagot2016improving}
Lo{\"\i}c Fagot-Bouquet, Romaric Audigier, Yoann Dhome, and Fr{\'e}d{\'e}ric
  Lerasle.
\newblock Improving multi-frame data association with sparse representations
  for robust near-online multi-object tracking.
\newblock In {\em European Conference on Computer Vision}, pages 774--790.
  Springer, 2016.

\bibitem{felzenszwalb2010object}
Pedro~F Felzenszwalb, Ross~B Girshick, David McAllester, and Deva Ramanan.
\newblock Object detection with discriminatively trained part-based models.
\newblock {\em IEEE transactions on pattern analysis and machine intelligence},
  32(9):1627--1645, 2010.

\bibitem{gardner1998artificial}
Matt~W Gardner and SR Dorling.
\newblock Artificial neural networks (the multilayer perceptron)—a review of
  applications in the atmospheric sciences.
\newblock {\em Atmospheric environment}, 32(14-15):2627--2636, 1998.

\bibitem{hamid2015joint}
Seyed Hamid~Rezatofighi, Anton Milan, Zhen Zhang, Qinfeng Shi, Anthony Dick,
  and Ian Reid.
\newblock Joint probabilistic data association revisited.
\newblock In {\em Proceedings of the IEEE international conference on computer
  vision}, pages 3047--3055, 2015.

\bibitem{he2017mask}
Kaiming He, Georgia Gkioxari, Piotr Doll{\'a}r, and Ross Girshick.
\newblock Mask r-cnn.
\newblock In {\em Proceedings of the IEEE international conference on computer
  vision}, pages 2961--2969, 2017.

\bibitem{he2016deep}
Kaiming He, Xiangyu Zhang, Shaoqing Ren, and Jian Sun.
\newblock Deep residual learning for image recognition.
\newblock In {\em Proceedings of the IEEE conference on computer vision and
  pattern recognition}, pages 770--778, 2016.

\bibitem{hermans2017defense}
Alexander Hermans, Lucas Beyer, and Bastian Leibe.
\newblock In defense of the triplet loss for person re-identification.
\newblock {\em arXiv preprint arXiv:1703.07737}, 2017.

\bibitem{huang2017densely}
Gao Huang, Zhuang Liu, Laurens Van Der~Maaten, and Kilian~Q Weinberger.
\newblock Densely connected convolutional networks.
\newblock In {\em Proceedings of the IEEE conference on computer vision and
  pattern recognition}, pages 4700--4708, 2017.

\bibitem{jiang2018online}
Na Jiang, SiChen Bai, Yue Xu, Chang Xing, Zhong Zhou, and Wei Wu.
\newblock Online inter-camera trajectory association exploiting person
  re-identification and camera topology.
\newblock In {\em 2018 ACM Multimedia Conference on Multimedia Conference},
  pages 1457--1465. ACM, 2018.

\bibitem{joo2007multiple}
Seong-Wook Joo and Rama Chellappa.
\newblock A multiple-hypothesis approach for multiobject visual tracking.
\newblock {\em IEEE Transactions on Image Processing}, 16(11):2849--2854, 2007.

\bibitem{kumar2014multiple}
Ratnesh Kumar, Guillaume Charpiat, and Monique Thonnat.
\newblock Multiple object tracking by efficient graph partitioning.
\newblock In {\em Asian Conference on Computer Vision}, pages 445--460.
  Springer, 2014.

\bibitem{leal2016learning}
Laura Leal-Taix{\'e}, Cristian Canton-Ferrer, and Konrad Schindler.
\newblock Learning by tracking: Siamese cnn for robust target association.
\newblock In {\em Proceedings of the IEEE Conference on Computer Vision and
  Pattern Recognition Workshops}, pages 33--40, 2016.

\bibitem{leal2015motchallenge}
Laura Leal-Taix{\'e}, Anton Milan, Ian Reid, Stefan Roth, and Konrad Schindler.
\newblock Motchallenge 2015: Towards a benchmark for multi-target tracking.
\newblock {\em arXiv preprint arXiv:1504.01942}, 2015.

\bibitem{leal2017tracking}
Laura Leal-Taix{\'e}, Anton Milan, Konrad Schindler, Daniel Cremers, Ian Reid,
  and Stefan Roth.
\newblock Tracking the trackers: an analysis of the state of the art in
  multiple object tracking.
\newblock {\em arXiv preprint arXiv:1704.02781}, 2017.

\bibitem{leibe2007coupled}
Bastian Leibe, Konrad Schindler, and Luc Van~Gool.
\newblock Coupled detection and trajectory estimation for multi-object
  tracking.
\newblock In {\em Computer Vision, 2007. ICCV 2007. IEEE 11th International
  Conference on}, pages 1--8. IEEE, 2007.

\bibitem{liu2017end}
Hao Liu, Jiashi Feng, Meibin Qi, Jianguo Jiang, and Shuicheng Yan.
\newblock End-to-end comparative attention networks for person
  re-identification.
\newblock {\em IEEE Transactions on Image Processing}, 26(7):3492--3506, 2017.

\bibitem{liu2016ssd}
Wei Liu, Dragomir Anguelov, Dumitru Erhan, Christian Szegedy, Scott Reed,
  Cheng-Yang Fu, and Alexander~C Berg.
\newblock Ssd: Single shot multibox detector.
\newblock In {\em European conference on computer vision}, pages 21--37.
  Springer, 2016.

\bibitem{maksai2017non}
Andrii Maksai, Xinchao Wang, Fran{\c{c}}ois Fleuret, and Pascal Fua.
\newblock Non-markovian globally consistent multi-object tracking.
\newblock In {\em 2017 IEEE International Conference on Computer Vision
  (ICCV)}, pages 2563--2573. IEEE, 2017.

\bibitem{milan2016mot16}
Anton Milan, Laura Leal-Taix{\'e}, Ian Reid, Stefan Roth, and Konrad Schindler.
\newblock Mot16: A benchmark for multi-object tracking.
\newblock {\em arXiv preprint arXiv:1603.00831}, 2016.

\bibitem{milan2014continuous}
Anton Milan, Stefan Roth, and Konrad Schindler.
\newblock Continuous energy minimization for multitarget tracking.
\newblock {\em IEEE Trans. Pattern Anal. Mach. Intell.}, 36(1):58--72, 2014.

\bibitem{ren2015faster}
Shaoqing Ren, Kaiming He, Ross Girshick, and Jian Sun.
\newblock Faster r-cnn: Towards real-time object detection with region proposal
  networks.
\newblock In {\em Advances in neural information processing systems}, pages
  91--99, 2015.

\bibitem{ristani2016performance}
Ergys Ristani, Francesco Solera, Roger Zou, Rita Cucchiara, and Carlo Tomasi.
\newblock Performance measures and a data set for multi-target, multi-camera
  tracking.
\newblock In {\em European Conference on Computer Vision}, pages 17--35.
  Springer, 2016.

\bibitem{ristani2018features}
Ergys Ristani and Carlo Tomasi.
\newblock Features for multi-target multi-camera tracking and
  re-identification.
\newblock {\em arXiv preprint arXiv:1803.10859}, 2018.

\bibitem{sadeghian2017tracking}
Amir Sadeghian, Alexandre Alahi, and Silvio Savarese.
\newblock Tracking the untrackable: Learning to track multiple cues with
  long-term dependencies.
\newblock In {\em Proceedings of the IEEE International Conference on Computer
  Vision}, pages 300--311, 2017.

\bibitem{schroff2015facenet}
Florian Schroff, Dmitry Kalenichenko, and James Philbin.
\newblock Facenet: A unified embedding for face recognition and clustering.
\newblock In {\em Proceedings of the IEEE conference on computer vision and
  pattern recognition}, pages 815--823, 2015.

\bibitem{shine2019comparative}
Linu Shine, Anitha Edison, and CV Jiji.
\newblock A comparative study of faster r-cnn models for anomaly detection in
  2019 ai city challenge.
\newblock In {\em Proceedings of the IEEE Conference on Computer Vision and
  Pattern Recognition Workshops}, pages 306--314, 2019.

\bibitem{shitrit2014multi}
Horesh~Ben Shitrit, J{\'e}r{\^o}me Berclaz, Fran{\c{c}}ois Fleuret, and Pascal
  Fua.
\newblock Multi-commodity network flow for tracking multiple people.
\newblock {\em IEEE transactions on pattern analysis and machine intelligence},
  36(8):1614--1627, 2014.

\bibitem{singh2008pedestrian}
Vivek~Kumar Singh, Bo Wu, and Ramakant Nevatia.
\newblock Pedestrian tracking by associating tracklets using detection
  residuals.
\newblock In {\em Motion and video Computing, 2008. WMVC 2008. IEEE Workshop
  on}, pages 1--8. IEEE, 2008.

\bibitem{sun2018PCB}
Yifan Sun, Liang Zheng, Yi Yang, Qi Tian, and Shengjin Wang.
\newblock Beyond part models: Person retrieval with refined part pooling (and a
  strong convolutional baseline).
\newblock In {\em ECCV}, 2018.

\bibitem{tang2019cityflow}
Zheng Tang, Milind Naphade, Ming-Yu Liu, Xiaodong Yang, Stan Birchfield, Shuo
  Wang, Ratnesh Kumar, David Anastasiu, and Jenq-Neng Hwang.
\newblock Cityflow: A city-scale benchmark for multi-target multi-camera
  vehicle tracking and re-identification.
\newblock In {\em Proceedings of the IEEE Conference on Computer Vision and
  Pattern Recognition}, pages 8797--8806, 2019.

\bibitem{tang2018single}
Zheng Tang, Gaoang Wang, Hao Xiao, Aotian Zheng, and Jenq-Neng Hwang.
\newblock Single-camera and inter-camera vehicle tracking and 3d speed
  estimation based on fusion of visual and semantic features.
\newblock In {\em Proceedings of the IEEE Conference on Computer Vision and
  Pattern Recognition Workshops}, pages 108--115, 2018.

\bibitem{tesfaye2017multi}
Yonatan~Tariku Tesfaye, Eyasu Zemene, Andrea Prati, Marcello Pelillo, and
  Mubarak Shah.
\newblock Multi-target tracking in multiple non-overlapping cameras using
  constrained dominant sets.
\newblock {\em arXiv preprint arXiv:1706.06196}, 2017.

\bibitem{thoreau2018improving}
Michael Thoreau and Navinda Kottege.
\newblock Improving online multiple object tracking with deep metric learning.
\newblock {\em arXiv preprint arXiv:1806.07592}, 2018.

\bibitem{varior2016gated}
Rahul~Rama Varior, Mrinal Haloi, and Gang Wang.
\newblock Gated siamese convolutional neural network architecture for human
  re-identification.
\newblock In {\em European Conference on Computer Vision}, pages 791--808.
  Springer, 2016.

\bibitem{wang2014tracklet}
Bing Wang, Gang Wang, Kap Luk~Chan, and Li Wang.
\newblock Tracklet association with online target-specific metric learning.
\newblock In {\em Proceedings of the IEEE Conference on Computer Vision and
  Pattern Recognition}, pages 1234--1241, 2014.

\bibitem{wang2016tracking}
Xinchao Wang, Engin T{\"u}retken, Francois Fleuret, and Pascal Fua.
\newblock Tracking interacting objects using intertwined flows.
\newblock {\em IEEE transactions on pattern analysis and machine intelligence},
  38(EPFL-ARTICLE-210040):2312--2326, 2016.

\bibitem{xiang2018multiple}
Jun Xiang, Guoshuai Zhang, Jianhua Hou, Nong Sang, and Rui Huang.
\newblock Multiple target tracking by learning feature representation and
  distance metric jointly.
\newblock {\em arXiv preprint arXiv:1802.03252}, 2018.

\bibitem{yoon2018multiple}
Kwangjin Yoon, Young-min Song, and Moongu Jeon.
\newblock Multiple hypothesis tracking algorithm for multi-target multi-camera
  tracking with disjoint views.
\newblock {\em IET Image Processing}, 12(7):1175--1184, 2018.

\bibitem{yu2016solution}
Shoou-I Yu, Deyu Meng, Wangmeng Zuo, and Alexander Hauptmann.
\newblock The solution path algorithm for identity-aware multi-object tracking.
\newblock In {\em Proceedings of the IEEE Conference on Computer Vision and
  Pattern Recognition}, pages 3871--3879, 2016.

\bibitem{zhang2017multi}
Zhimeng Zhang, Jianan Wu, Xuan Zhang, and Chi Zhang.
\newblock Multi-target, multi-camera tracking by hierarchical clustering:
  Recent progress on dukemtmc project.
\newblock {\em arXiv preprint arXiv:1712.09531}, 2017.

\bibitem{zheng2015scalable}
Liang Zheng, Liyue Shen, Lu Tian, Shengjin Wang, Jingdong Wang, and Qi Tian.
\newblock Scalable person re-identification: A benchmark.
\newblock In {\em Proceedings of the IEEE International Conference on Computer
  Vision}, pages 1116--1124, 2015.

\bibitem{zheng2016person}
Liang Zheng, Yi Yang, and Alexander~G Hauptmann.
\newblock Person re-identification: Past, present and future.
\newblock {\em arXiv preprint arXiv:1610.02984}, 2016.

\bibitem{zheng2017unlabeled}
Zhedong Zheng, Liang Zheng, and Yi Yang.
\newblock Unlabeled samples generated by gan improve the person
  re-identification baseline in vitro.
\newblock In {\em Proceedings of the IEEE International Conference on Computer
  Vision}, 2017.

\bibitem{zhong2017random}
Zhun Zhong, Liang Zheng, Guoliang Kang, Shaozi Li, and Yi Yang.
\newblock Random erasing data augmentation.
\newblock {\em arXiv preprint arXiv:1708.04896}, 2017.

\end{thebibliography}
}

\end{document}